\documentclass[twoside,11pt]{article}

\usepackage{jmlr2e}

\usepackage{microtype}
\usepackage{graphicx}
\usepackage{booktabs} 

\usepackage{tabularx}
\usepackage{mystyle}
\usepackage{subcaption}
\usepackage{hyperref}
\usepackage{multirow}

 \usepackage{lastpage}
  \jmlrheading{21}{2020}{1-\pageref{LastPage}}{7/18}{7/20}{18-503}{Jean Kossaifi, Zachary C. Lipton, Arinbj\"orn Kolbeinsson, Aran Khanna, Tommaso Furlanello and Anima Anandkumar}

\ShortHeadings{Tensor Regression Networks}{Kossaifi, Lipton, Kolbeinsson, Khanna, Furlanello and Anandkumar}

\begin{document}

\title{Tensor Regression Networks}

\author{\name Jean Kossaifi \email jean.kossaifi@imperial.ac.uk \\
       \addr NVIDIA \& Imperial College London
       \AND
       \name Zachary C. Lipton \email zlipton@cmu.edu \\
       \addr Carnegie Mellon University
       \AND
       \name Arinbj\"orn Kolbeinsson \email ak711@imperial.ac.uk\\
       \addr Imperial College London
       \AND
       \name Aran Khanna \email arankhan@amazon.com \\
       \addr Amazon AI
       \AND
       \name Tommaso Furlanello \email furlanel@usc.edu\\
       \addr University of Southern California
       \AND
       \name Anima Anandkumar \email anima@caltech.edu\\
       \addr NVIDIA \&
       California Institute of Technology
}

\editor{Aapo Hyvarinen}
 
\maketitle

\begin{abstract}
Convolutional neural networks 
typically consist of many convolutional layers
followed by one or more fully connected layers. 
While convolutional layers map between 
high-order activation tensors, 
the fully connected layers 
operate on flattened activation vectors.
Despite empirical success,
this approach has notable drawbacks.
Flattening followed by fully connected layers 
discards multilinear structure in the activations 
and requires many parameters. 
We address these problems
by incorporating tensor algebraic operations 
that preserve multilinear structure at every layer.
First, we introduce \emph{Tensor Contraction Layers} (TCLs)
that reduce the dimensionality of their input 
while preserving their multilinear structure using tensor contraction. 
Next, we introduce \emph{Tensor Regression Layers} (TRLs),
which express outputs through a low-rank multilinear mapping 
from a high-order activation tensor 
to an output tensor of arbitrary order. 
We learn the contraction and regression factors end-to-end, 
and produce accurate nets with fewer parameters.
Additionally, our layers regularize networks
by imposing low-rank constraints 
on the activations (TCL) and regression weights (TRL).
Experiments on ImageNet show that, 
applied to VGG and ResNet architectures,
TCLs and TRLs reduce the number of parameters
compared to fully connected layers by more than 65\% 
while maintaining or increasing accuracy. 
In addition to the space savings, 
our approach's ability 
to leverage topological structure 
can be crucial for structured data such as MRI. 
In particular, we demonstrate significant performance improvements 
over comparable architectures 
on three tasks associated with the UK Biobank dataset. 
\end{abstract}

\begin{keywords}
    Machine Learning, Tensor Methods, Tensor Regression Networks, Low-Rank Regression, Tensor Regression Layers, Deep Learning, Tensor Contraction
\end{keywords}

\section{Introduction}
Many natural datasets exhibit multi-modal structure. 
We represent audio spectrograms as 2\mynd-order tensors (matrices)
with modes corresponding 
to frequency and time. 
We represent images as 3\myrd-order tensors 
with modes corresponding to  
width, height and the color channels.
Videos are expressed 
as 4\myth-order tensors,
and the signal processed 
by an array of video sensors 
can be described as a 5\myth-order tensor.
Multilinear structure arises naturally in many medical applications:
MRI images are 3\myrd-order tensors
and functional MRI images are 4\myth-order tensors.
Generally, a broad array of multimodal data
can be naturally encoded as tensors. 
Tensor methods extend linear algebra 
to higher order tensors and
are promising tools for manipulating and analyzing such data.
 
The mathematical properties of tensors
have long been the subject of theoretical study.
Previously, in machine learning,  
data points were typically assumed to be vectors 
and datasets to be matrices. 
Hence, spectral methods, 
such as matrix decompositions, 
have been popular in machine learning. 
Recently, tensor methods,
which generalize these techniques to higher-order tensors, have gained prominence.
One class of broadly useful techniques 
within tensor methods are tensor decompositions, 
which have been studied for a variety of applications
in signal processing and machine learning~\citep{tensor_decomposition_sidiropoulos,cichocki_2015}, 
data mining and fusion~\citep{tensor_mining_papalexakis}, 
blind source separation~\citep{nonnegative_cichocki_2009},
computer vision~\citep{tensor_faces}
and learning latent variable models~\citep{anandkumar2014tensor}.

Deep Neural Networks (DNNs) frequently 
manipulate high-order tensors:
in a standard deep Convolutional Neural Network (CNN) for image recognition,
the inputs and the activations 
of convolutional layers
are 3\myrd-order tensors. 
And yet, to wit, most architectures 
output predictions
by first flattening the activation tensors
and then connecting to the output neurons 
via one or more fully connected layers. 
This approach presents several issues: 
(i) we lose multimodal information
during the flattening process; 
and (ii) the fully connected layers
require a large number of parameters.

\paragraph{In this paper,}
we propose Tensor Contraction Layers (TCLs) 
and Tensor Regression Layers (TRLs)
as end-to-end trainable 
components of neural networks. 
In doing so, we exploit multilinear structure 
without giving up the power and flexibility 
offered by modern deep learning methods. 
By replacing fully connected layers 
with tensor contractions, 
we aggregate long-range spatial information 
while preserving multi-modal structure.
Moreover, by enforcing low rank, 
we reduce the number of parameters needed significantly with minimal impact on accuracy.

Our proposed TRL
expresses the regression weights
through the factors of a low-rank tensor decomposition. 
The TRL obviates the need for flattening,
instead leveraging the structure
when generating output.
By combining tensor regression 
with tensor contraction,
we further increase efficiency. 
Augmenting the VGG and ResNet architectures, 
we demonstrate improved performance on the ImageNet dataset 
despite significantly reducing the number of parameters (almost by 65\%). 
The ability to preserve the topological structure in the data 
is particularly crucial for prediction from MRI data. 
In this context, we conduct experiments for $3$ different tasks 
(gender classification, body mass index prediction and age regression) on the UK biobank dataset, 
the largest available MRI dataset. 
There, we demonstrate superior performance with our approach
and show large performance improvements for all $3$ tasks.
This is the first paper to present 
an end-to-end trainable architecture 
that retains the multi-modal tensor structure throughout the network.

\paragraph{ Related work: }
Several recent papers apply tensor decomposition to deep learning. 
One notable line of application is to re-parametrize existing layers
using tensor decomposition either to speed these up or reduce the number of parameters.
\citet{lebedev2014speeding} propose using CP decomposition 
to speed up convolutional layers. 
Similarly, \citep{tai2015convolutional} propose to use tensor decomposition 
to remove redundancy in convolutional layers and express these 
as the composition of two convolutional layers with less parameters. 

\citet{yong2015compression} take a pre-trained network 
and apply tensor (Tucker) decomposition on the convolutional kernel tensors
and then fine-tune the resulting network. 
\citet{yang2016deep} propose weight sharing 
in multitask learning and \citet{chen2017sharing} 
propose sharing residual units.
\citet{novikov2015tensorizing} use 
the Tensor-Train (TT) format  
to impose low-rank tensor structure on weights 
of the fully connected layers in order to compress them. 
However, they still retain 
the fully connected layers for the output.
In addition, the reshaping to arbitrary higher orders and dimensions
does not guarantee that the multilinear structure is preserved. 
By contrast, we present an end-to-end tensorized network architecture 
that focuses on leveraging that structure.
Many of these contributions are orthogonal to ours and can be applied together.

Despite the success of DNNs, 
many open questions remain
as to why they work so well
and whether they really need so many parameters.
Tensor methods have emerged as promising tools of analysis 
to address these questions
and to better understand the success of deep neural networks.
\citet{cohen2015expressive}, for example,
use tensor methods as tools of analysis 
to study the expressive power of CNNs, 
while the follow up work \citep{sharir2017expressive} 
focuses on the expressive power of overlapping architectures of deep learning. 
\citet{haeffele2015global} derive sufficient conditions
for global optimality and optimization of non-convex factorization problems,
including tensor factorization and deep neural network training.
\citet{Grefenstette2011experimental} explore 
contraction as a composition operator for NLP.
Other papers investigate tensor methods 
as tools for devising neural network learning algorithms
with theoretical guarantees of convergence \citep{sedghi2016training,janzamin2015generalization,non_convexity_nn_anandkumar}.

Several prior papers address the 
power of tensor regression to
preserve natural multi-modal structure 
and learn compact predictive models~\citep{guo2012tensor,rabusseau2016low,zhou2013tensor,yu2016learning}.
However, these works typically rely on analytical solutions 
and require manipulating large tensors containing the data.
They are usually used for small datasets or require to downsample the data 
or extract compact features prior to fitting the model,
and do not scale to large datasets such as ImageNet.

To our knowledge, no prior work combines tensor contraction or tensor regression 
with deep learning in an end-to-end trainable fashion.

\section{Mathematical background}
\paragraph{Notation}
Throughout the paper, we define tensors as multidimensional arrays, 
with indexing starting at 0. 
First order tensors are vectors, denoted \(\myvector{v}\). 
Second order tensors are matrices, denoted \(\mymatrix{M}\) 
and \(\myId\) is the identity matrix. 
By \(\mytensor{X}\), we denote tensors of order 3 or greater. 
For a third order tensor \(\mytensor{X}\),
we denote its element \((i, j, k)\) as \(\mytensor{X}_{i_1, i_2, i_3}\). 
A colon is used to denote all elements of a mode, e.g.,
the mode-1 fibers of \(\mytensor{X}\) 
are denoted as \(\mytensor{X}_{:, i_2, i_3}\). 
The transpose of \(\mymatrix{M}\) is denoted \(\mymatrix{M}\myT\). 
Finally, for any \(i, j \in \myN,  i < j, \myrange{i}{j}\) denotes
the set of integers \(\{ i, i+1, \cdots , j-1, j\}\). 

\paragraph{Tensor unfolding}
Given a tensor,
\( \mytensor{X} \in \myR^{I_0 \times I_1 \times \cdots \times I_N}\),
its mode-\(n\) unfolding is a matrix \(\mymatrix{X}_{[n]} \in \myR^{I_n \times I_M}\), 
with \(I_M = \prod_{\substack{k=0,\\k \neq n}}^N I_k\)
and is defined by the mapping from element
\( (i_0, i_1, \cdots, i_N)\) to \((i_n, j)\), with 
\(
j = \sum_{\substack{k=0,\\k \neq n}}^N i_k \times \prod_{\substack{l=k+1,\\ l \neq n}}^N I_l 
\). 
We use the definition introduced in~\citet{tensorly},
which corresponds to an underlying row-wise ordering of the elements. 
This differs from the definition used by~\citet{kolda2009tensor},
which correponds to an underlying column-wise ordering of the elements. Throughout the paper, we assume the elements are arranged in a row-wise manner, which is reflected in the definitions of unfolding, 
vectorization, and the resulting formulas. 
This row-ordering elements matches the actual ordering on GPUs
and allows for more efficient implementation.

\paragraph{Tensor vectorization}
Given a tensor,
\( \mytensor{X} \in \myR^{I_0 \times I_1 \times \cdots \times I_N}\), 
we can flatten it into a vector \(\text{vec}(\mytensor{X})\) 
of size \(\left(I_0 \times \cdots \times I_N\right)\) 
defined by the mapping from element
\( (i_0, i_1, \cdots, i_N)\) of \(\mytensor{X}\) to element \(j\) of \(\text{vec}(\mytensor{X})\), with 
\( j = \sum_{k=0}^N i_k \times \prod_{m=k+1}^N I_m \). As for unfolding, we assume a row-ordering of the elements, following~\citet{tensorly}.

\paragraph{n-mode product}
For a tensor \(\mytensor{X} \in \myR^{I_0 \times I_1 \times \cdots \times I_N}\) and a matrix \( \mymatrix{M} \in \myR^{R \times I_n} \), the n-mode product of a tensor 
is a tensor of size 
\(\left(I_0 \times \cdots \times I_{n-1} \times R \times I_{n+1} \times \cdot \times I_N\right)\) 
and can be expressed using unfolding of \(\mytensor{X}\) 
and the classical dot product as
\begin{equation}
	\mytensor{X} \times_n \mymatrix{M} = \mymatrix{M} \mytensor{X}_{[n]} \in \myR^{I_0 \times \cdots \times I_{n-1} \times R \times I_{n+1} \times \cdots \times I_N}.
\end{equation}

\paragraph{Generalized inner-product}
For two tensors \(\mytensor{X}, \mytensor{Y} \in \myR^{I_0 \times I_1 \times \cdots \times I_N}\) of same size, their inner product is defined as 
\(
\myinner{\mytensor{X}}{\mytensor{Y}}=~\sum_{i_0=0}^{I_0-1}\sum_{i_1=0}^{I_1 - 1} \cdots \sum_{i_n=0}^{I_N - 1} \mytensor{X}_{i_0, i_1, \cdots, i_n} \mytensor{Y}_{i_0, i_1, \cdots, i_n}
\) 
For two tensors \(\mytensor{X} \in \myR^{I_x \times I_1 \times I_2 \times \cdots \times I_N}\) and \( \mytensor{Y} \in \myR^{I_1 \times I_2 \times \cdots \times I_N \times I_y} \) sharing \(N\) modes of same size,
we similarly define a ``generalized inner product'' 
along the \(N\) last (respectively first) modes 
of \(\mytensor{X}\) (respectively \(\mytensor{Y}\)) as 
\begin{equation}
\myinner{\mytensor{X}}{\mytensor{Y}}_N =~\sum_{i_1=0}^{I_1-1}\sum_{i_2=0}^{I_1 - 1} \cdots \sum_{i_n=0}^{I_N - 1} \mytensor{X}_{:, i_1, i_2, \cdots, i_n} \mytensor{Y}_{i_1, i_2, \cdots, i_n, :},
\end{equation}
with \( \myinner{\mytensor{X}}{\mytensor{Y}}_N \in \myR^{I_x \times I_y}\).

\paragraph{Tucker decomposition}
Given a tensor \(\mytensor{X} \in \myR^{I_0 \times I_1 \times \cdots \times I_N} \), 
we can decompose it into a low rank core 
\(\mytensor{G} \in \myR^{R_0 \times R_1 \times \cdots \times R_N}\) 
by projecting along each of its modes
with projection factors 
\( \left( \mymatrix{U}^{(0)}, \cdots,\mymatrix{U}^{(N)} \right) \), with \(\mymatrix{U}^{(k)} \in \myR^{R_k \times I_k}, k \in \myrange{0}{N}\).
In other words, we can write
\begin{align}
\mytensor{X} &= 
\mytensor{G} \times_0 \mymatrix{U}^{(0)} 
		  \times_1  \mymatrix{U}^{(2)} \times
		  \cdots
          \times_N \mymatrix{U}^{(N)} \nonumber \\
        &= \mytucker{\mytensor{G}}{\mymatrix{U}^{(0)},
		  \cdots,
          \mymatrix{U}^{(N)}}.
\end{align}
Typically, the factors and core of the decomposition 
are obtained by solving a least squares problem. 
In particular, closed form solutions 
can be obtained for the factor by considering the \(n-\)mode unfolding of \(\mytensor{X}\) 
that can be expressed as
\begin{equation}
  \mymatrix{X}_{[n]} = \mymatrix{U}^{(n)} \mymatrix{G}_{[n]}
    				   \left(\mymatrix{U}^{(-k)}
                       \right)\myT,
\label{eq:unfold_tucker}
\end{equation}
where \(\mymatrix{U}^{(-k)}\) is defined as follows:
\begin{equation}
\mymatrix{U}^{(-k)}  =  \mymatrix{U}^{(0)}
                       \otimes \cdots 
                       \mymatrix{U}^{(n-1)} 
                       \otimes \mymatrix{U}^{(n+1)}
                       \otimes \cdots
                       \otimes \mymatrix{U}^{(N)}.\nonumber
\end{equation}
Notice the natural ordering of the factors, from \(0\) to \(N\), 
that follows from our definition of unfolding. 

Similarly, we can optimize the core in a straightforward manner 
by isolating it using the equivalent rewriting of the above equality:
\begin{equation}
  \myvec(\mymatrix{X}) = 
  \left( \mymatrix{U}^{(0)} \otimes \cdots \otimes \mymatrix{U}^{(N)} \right) \myvec(\mymatrix{G}).
\end{equation}
We refer the interested reader to the thorough review of the literature 
on tensor decompositions by \citet{kolda2009tensor}.
 
\section{Tensor Contraction Layer}
One natural way to incorporate 
tensor operations into a neural network 
is to apply tensor contraction 
to an activation tensor in order to obtain 
a low-dimensional representation~\citep{TCL}. In this section, we explain how to incorporate tensor contractions
into neural networks as a differentiable layer.

We call this technique the Tensor Contraction layer (TCL).
Compared to performing a similar rank reduction
with a fully connected layer,
TCLs require fewer parameters
and less computation, while preserving the multilinear structure of the activation tensor.

\begin{figure}[ht]
  \centering
  \includegraphics[width=0.5\linewidth]{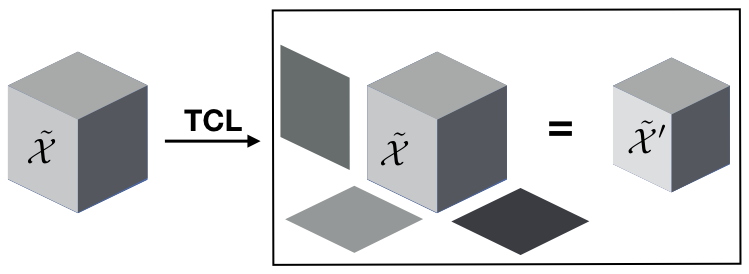}
  \caption{A representation of the Tensor Contraction Layer (TCL) on a tensor of order 3.
The input tensor \(\mytensor{X}\) is contracted into a low rank core \(\mytensor{X'}\). In practice, for efficiency, tensor contraction is applied to a \emph{mini-batch} of such activation tensors, not just a single sample as represented here.}\label{TCL_visual}
\end{figure}

\subsection{Tensor contraction layers}

Given an activation tensor \(\mytensor{X}\) of size \( \left( S_0, I_0, I_1, \cdots, I_N \right) \), 
the TCL will produce a compact core tensor \(\mytensor{G}\) of smaller size \( \left( S_0, R_0, R_1, \cdots, R_N \right) \) defined as
\begin{equation}
\mytensor{X'} = 
\mytensor{X} \times_1 \mymatrix{V}^{(0)} 
		  \times_2  \mymatrix{V}^{(1)} \times
		  \cdots
          \times_{N+1} \mymatrix{V}^{(N)},
\end{equation}
with \(\mymatrix{V}^{(k)} \in \myR^{R_k \times I_k}, k \in \myrange{0}{N}\). 
Note that the projections start at the second mode
because the first mode \(S_0\) corresponds to the batch.

This layer allows us to contract the input activation tensor
without discarding its multi-linear structure. 
By contrast, a flattening layer followed 
by a fully connected layer would discard that information.
One way to see this is to recognise that a fully connected layer 
is simply a tensor contraction over the second mode of a batch 
of flattened activations, as exposed in Subsection~\ref{trl_fc} below.

The projection factors \( \left(\mymatrix{V}^{(k)}\right)_{k \in [1, \cdots N]}\)  are learned end-to-end with the rest of the network by gradient backpropagation. In the rest of this paper, we denote \emph{size--\(\left(R_0, \cdots, R_N\right)\) TCL}, or \emph{TCL--\(\left(R_0, \cdots, R_N\right)\)}
a TCL that produces a compact core of dimension \(\left(R_0, \cdots, R_N\right)\).

\subsection{Gradient back-propagation}
In the case of the TCL, 
we simply need to take the gradients with respect to the factors \(\mymatrix{V}^{(k)}\) for each \(k \in {0, \cdots, N}\) of the tensor contraction. Specifically, we compute
\begin{equation}\nonumber
\myd{\mytensor{X'}}{\mymatrix{V}^{(k)}} = 
\myd{\mytensor{X} \times_1 \mymatrix{V}^{(0)} 
		  \times_2  \mymatrix{V}^{(1)} \times
		  \cdots
          \times_{N+1} \mymatrix{V}^{(N)}}{\mymatrix{V}^{(k)}}.
\end{equation}
By rewriting the previous equality in terms of unfolded tensors,
we get an equivalent rewriting where we have isolated the considered factor:
\begin{equation}\nonumber
\myd{\mytensor{X'}_{[k]}}{\mymatrix{V}^{(k)}} = 
 \myd{\mymatrix{V}^{(k)} \mymatrix{X}_{[k]}
    				   \left(\myId \otimes
                             \mymatrix{V}^{(-k)}
                       \right)\myT
             }{\mymatrix{V}^{(k)}},
\label{eq:tcl_layer_gradient}
\end{equation}
with 
\begin{equation}\nonumber
\mymatrix{V}^{(-k)}   =	\mymatrix{V}^{(0)}
                       \otimes \cdots 
                       \mymatrix{V}^{(k-1)} 
                       \otimes \mymatrix{V}^{(k+1)}
                       \otimes \cdots
                       \otimes \mymatrix{V}^{(N)}.
\end{equation}

\begin{figure*}[ht]
    \centering
    \vspace*{0.4cm}
    \includegraphics[width=1\linewidth]{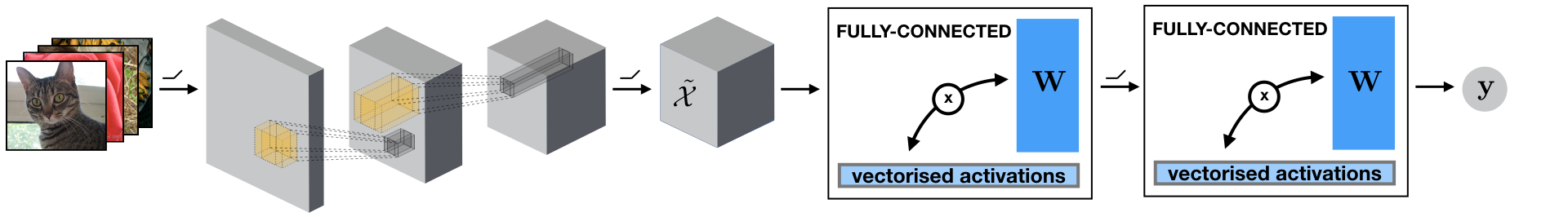}~\vspace{0.1cm}
    \caption{In standard CNNs,  the input \(\mytensor{X}\) is flattened  and then passed to a fully connected layer, where it is multiplied by a weight matrix \(\mymatrix{W}\).}
\end{figure*}

\begin{figure*}[ht]
    \centering
    \includegraphics[width=1\linewidth]{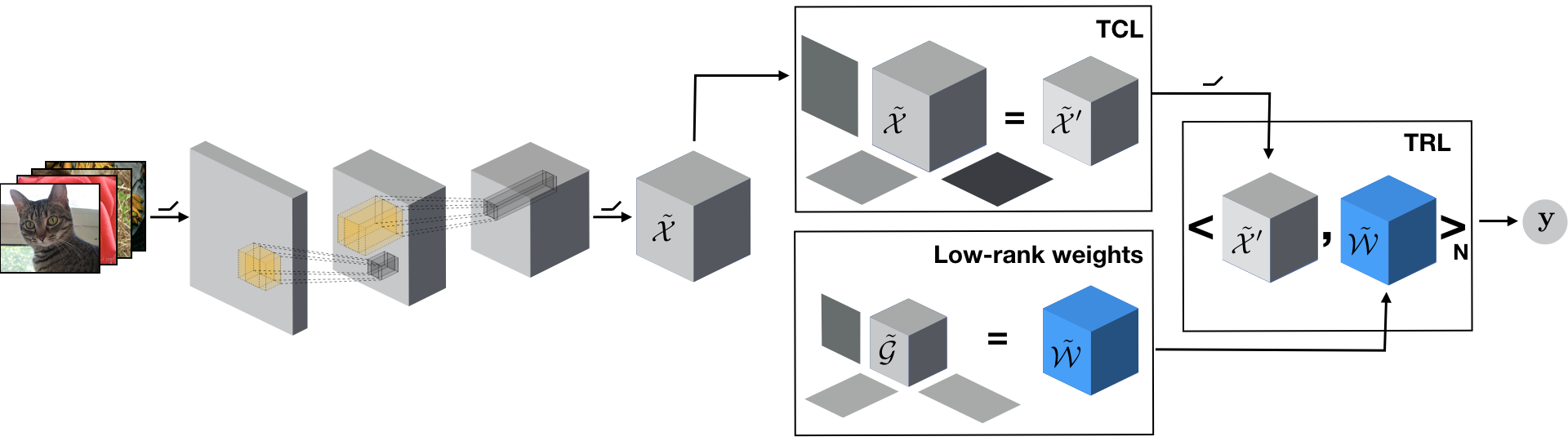}
    \caption{We propose to first reduce the dimensionality of the activation tensor by applying tensor contraction before performing tensor regression.
    We then replace flattening operators and fully connected layers 
 by a TRL.  The output is a product between the activation tensor and a low-rank weight tensor \(\mytensor{W}\). For clarity, we illustrate the case of a binary classification, where \(y\) is a scalar. For multiclass, \(y\) becomes a vector and the regression weights would become a \(4^{th}\) order tensor.}
\end{figure*}

\subsection{Model analysis}\label{trl_fc}

\paragraph{Link with fully connected layers} 
Let's considering an activation tensor \(\mytensor{X}\) of size \( \left( S_0, I_0, I_1, \cdots, I_N \right)\).
A size--\( \left(R_0, R_1, \cdots, R_N \right) \) TCL 
parameterized by weight factors \(\mymatrix{V}^{(0)}, \cdots, \mymatrix{V}^{(N)}\) 
is equivalent to a fully connected layer 
parametrized by the weight matrix \( \mymatrix{W} = \left(\mymatrix{V}^{(0)}  \otimes \cdots \otimes \mymatrix{V}^{(N)} \right)\myT \) and would compute
\[
    \mytensor{X}_{[0]} \mymatrix{W} =  \mytensor{X}_{[0]}  \left(\mymatrix{V}^{(0)}  \otimes \cdots \otimes \mymatrix{V}^{(N)} \right)\myT,
\]
where \(\mytensor{X}_{[0]}\) corresponds to the unfolding of \mytensor{X} 
along the first mode, e.g., a vectorization of each of the samples in the batch.

\paragraph{Number of parameters}
Considering an activation tensor \(\mytensor{X}\) of size
\( \left( S_0, I_0, I_1, \cdots, I_N \right) \), 
a size--\( \left(R_0, R_1, \cdots, R_N \right) \) TCL parameterized by weight factors \(\mymatrix{V}^{(0)}, \cdots, \mymatrix{V}^{(N)}\) and
taking \(\mytensor{X}\) as input will have 
a total of \( \sum_{k=0}^{N} I_k \times R_k \) parameters.

This is to contrast with an equivalent fully connected layer (as presented above), parametrized a weight matrix \( \mymatrix{W} = \left(\mymatrix{V}^{(0)}  \otimes \cdots \otimes \mymatrix{V}^{(N)} \right)\myT \), which would have a total of \( \prod{k=0}^{N} I_k \times R_k \) parameters.

Notice how the product in number of parameters of the fully connected layer becomes a sum when using a TCL. In other words, in addition to preserving the topological structure in the activation tensor, the TCL has significantly less parameters than a corresponsing fully connected layer.

\section{Tensor Regression Layer}

In this section, we introduce the Tensor Regression Layer, 
a new differentiable neural network layer.
In order to generate outputs, 
CNNs typically either flatten the activations 
or apply a spatial pooling operation. 
In either case, they discard all multimodal structure
and subsequently apply a fully-connected output layer.
Instead, we propose leveraging that multilinear structure 
in the activation tensor 
and formulate the output as lying in a low-rank subspace 
that jointly models the input and the output. 
We do this by means of a low-rank tensor regression, 
where we enforce a low multilinear rank 
of the regression weight tensor.

\subsection{Tensor regression as a layer}
Let us denote by
\(\mytensor{X} \in \myR^{S \times I_0 \times I_1 \times \cdots \times I_N} \)
the input activation tensor corresponding to a batch of \(S\) samples \(\left(\mytensor{X}_1, \cdots, \mytensor{X}_S\right)\) and 
\( \mymatrix{Y} \in
\myR^{ S \times O } 
\)
the \(O\) corresponding labels for each sample. 
We are interested in the problem 
of estimating the regression weight tensor 
\( \mytensor{W} \in 
   \myR^{I_0 \times I_1 \times \cdots \times I_N \times O}
\)
under some fixed low rank \( \left(R_0, \cdots, R_N, R_{N+1}\right)\) and a bias \(\myvector{b} \in \myR^{O}\), such that \( \mymatrix{Y} = \myinner{\mytensor{X}}{\mytensor{W}}_N + \myvector{b}\), i.e.,
\begin{align}\label{eq:trl}
    \myvector{Y} & = \myinner{\mytensor{X}}{\mytensor{W}}_N + \myvector{b} \nonumber \\
    \text{subject to } & \mytensor{W} =  
    \mytucker{\mytensor{G}}{\mymatrix{U}^{(0)},
		  \cdots,
          \mymatrix{U}^{(N)}, \mymatrix{U}^{(N + 1)}},
\end{align}
with \(\myinner{\mytensor{X}}{\mytensor{W}}_N = \mytensor{X}_{[0]} \times {\mytensor{W}}_{[N+1]}\) the contraction of \(\mytensor{X}\) by \( \mytensor{W}\) along their \(N\) last (respectively first) modes, \(\mytensor{G} \in \myR^{R_0 \times \cdots \times R_N \times R_{N+1}} \), 
\(\mymatrix{U}^{(k)} \in \myR^{I_k \times R_k}\) for each \(k\) in \(\myrange{0}{N}\)
and \(\mymatrix{U}^{(N+1)} \in \myR^{O \times R_{N+1}}\).
 
Previously, this setting has been studied as a standalone problem. 
In that setting, the input data is directly mapped to the output,
and the problem solved analytically
\citep{guo2012tensor,rabusseau2016low,zhou2013tensor,yu2016learning}.
However, this either limits the model to raw data 
(e.g., pixel intensities) or requires pre-processing the data 
to extract (hand-crafted) features to feed the model.
For instance, fiducial points that encode the geometry (e.g. facial landmarks) are extracted~\citep{guo2012tensor}.
In addition, analytical solutions are prohibitive 
in terms of computation and memory usage for large datasets.

In this work, we incorporate tensor regressions as trainable layers in neural networks, which allow to learn jointly the features (e.g. via convolutional layers) and the tensor regression.
We do so by replacing the traditional 
flattening + fully connected layers 
with a tensor regression applied directly 
to the high-order input and enforcing 
low rank constraints on the weights of the regression. 
We call our layer the \emph{Tensor Regression Layer} (\emph{TRL}). 
Intuitively, the advantage of the TRL comes from leveraging the multi-modal structure in the data and expressing the solution as lying on a low rank manifold encompassing both the data and the associated outputs.

\subsection{Gradient backpropagation}

The gradients of the regression weights and the core 
with respect to each factor 
can be obtained by writing:
\begin{equation}
\myd{\mytensor{W}}{\mymatrix{U}^{(k)}}=
\myd{\mytensor{G} \times_0 \mymatrix{U}^{(0)} 
		  \times_1  \mymatrix{U}^{(1)} \times
		  \cdots
          \times_{N+1} \mymatrix{U}^{(N+1)}}{\mymatrix{U}^{(k)}}\nonumber
\end{equation}
Using the unfolded expression of the regression weights, we obtain the equivalent formulation:
\begin{equation}
\myd{\mytensor{W}_{[k]}}{\mymatrix{U}^{(k)}}=
\myd{\mymatrix{U}^{(k)} \mymatrix{G}_{[k]}
    				    \mathbf{R}\myT
             }{\mymatrix{U}^{(k)}},
\label{eq:layer_gradient}
\end{equation}
with
\begin{equation}\nonumber
\mathbf{R} = \mymatrix{U}^{(0)}
                       \otimes \cdots 
                       \mymatrix{U}^{(k-1)} 
                       \otimes \mymatrix{U}^{(k+1)}
                       \otimes \cdots
                       \otimes \mymatrix{U}^{(N+1)}. 
\end{equation}
Similarly, we can obtain the gradient with respect to the core 
by considering the vectorized expressions:
\begin{equation}\nonumber
  \myd{\text{vec}(\mytensor{W})}{\text{vec}(\mytensor{G})}=
  \myd{\left( \mymatrix{U}^{(0)} \otimes \cdots \otimes \mymatrix{U}^{(N+1)} \right) vec(\mymatrix{G})}{
  \text{vec}(\mytensor{G})}.
\end{equation}

\subsection{Model analysis}
We consider as input an activation tensor \(\mytensor{X} \in \myR^{S \times I_0 \times I_1 \times \cdots \times I_N} \), 
and a rank-\((R_0, R_1, \cdots, R_N, R_{N+1})\) tensor regression layer, 
where, typically, \(R_k \leq I_k\). 
Let's assume the output is \(n\)-dimensional.
A fully connected layer taking \(\mytensor{X}\) (after a flattening layer) as input will have \(n_{\text{FC}}\) parameters, with
\begin{equation}\nonumber
n_{\text{FC}} = n \times \prod_{k=0}^N I_k
\end{equation}
By comparison, a rank-\((R_0, R_1, \cdots, R_N, R_{N+1})\) TRL taking \(\mytensor{X}\) as input has a number of parameters \(n_{\text{TRL}}\), with:
\begin{equation}\nonumber
n_{\text{TRL}} =  \prod_{k=0}^{N+1} R_k + \sum_{k=0}^N R_k \times I_k + R_{N+1} \times n.
\end{equation}

\section{Efficient implementation of tensor regression layers}

Based on the previous layers, we propose an equivalent,
more efficient practical implementation of the tensor regression layer.
Equation~\ref{eq:trl} can be written:
\begin{equation}\nonumber
    \mymatrix{Y} = \myinner
        {\mytensor{X}}
        {\mytensor{G} \times_0 \mymatrix{U}^{(0)}  \times_1  \mymatrix{U}^{(1)} \times \cdots  \times_{N+1} \mymatrix{U}^{(N+1)} }_N + \myvector{b}
\end{equation}
We can rewrite this equivalently as
\begin{equation}\nonumber
    \myvector{Y} = \myinner
        {\mytensor{X} \times_0 (\mymatrix{U}^{(0)})\myT  \times_1  (\mymatrix{U}^{(1)})\myT \times \cdots  \times_{N}(\mymatrix{U}^{(N)})\myT }
        { \mytensor{G}  \times_{N+1} \mymatrix{U}^{(N+1)} }_N + \myvector{b}.
\end{equation}
This way, most of the computation is done in the \emph{low-rank} subspace 
rather than directly on the dimensions of \(\mytensor{X}\).

In practice, we use this formulation for implementation, 
and directly learn the pseudo-inverse of each factor\(\mymatrix{V}^{(k)} \in \myR^{I_k \times R_k}\) for each \(k\) in \(\myrange{0}{N}\):
\begin{equation}\nonumber
    \myvector{Y} = \myinner
        {\mytensor{X} \times_0 \mymatrix{V}^{(0)}  \times_1  \mymatrix{V}^{(1)} \times \cdots  \times_{N} \mymatrix{V}^{(N)}}
        { \mytensor{G}  \times_{N+1} \mymatrix{U}^{(N+1)} }_N + \myvector{b}
\end{equation}
Note that this is equivalent to first applying a TCL on \mytensor{X},
and then applying a TRL on the result to produce \myvector{Y},
with the first factors set to the identity 
(i.e., the low-rank constraint is applied only 
to the modes corresponding to the output).

In addition to this efficient formulation, 
it is also possible to achieve computational speedup via hardware acceleration 
by leveraging recent work, e.g., \citet{shi2016tensor},
on extending BLAS primitives. 
This would allow us to reduce the computational overhead for transpositions,
which are necessary when computing tensor contractions.
 
\section{Experiments}
We empirically demonstrate the effectiveness 
of preserving the tensor structure 
through tensor contraction and tensor regression 
by integrating it into state-of-the-art architectures 
and demonstrating similar performance 
on the popular ImageNet dataset.
We show that a TRL gives equal or greater performance 
as compared to flattening followed by fully connected layers,
while allowing for large space savings.
In particular, we show that our proposed layers allow us 
to best leverage multi-linear structure in data. 
This is particularly important for MRI data. 
On the largest database available, the U.K. biobank, 
we show that this approach outperforms its traditional counterparts
by large margins on three separate prediction tasks.

We empirically verify the effectiveness of the TCL on VGG-19~\citep{vgg}. 
We also conduct thorough experiments and ablation studies 
with the proposed layers on VGG-19, AlexNet, ResNet-50 and ResNet-101~\citep{resnet} in various scenarios.

\subsection{Implementation details}
We implemented all models using the MXNet library~\citep{mxnet} 
as well as the PyTorch library~\cite{paszke2017automatic}. For all tensor methods, we used the TensorLy library~\cite{tensorly}s.
The models were trained with data parallelism 
across multiple GPUs on Amazon Web Services, 
with 4 NVIDIA k80 GPUs. 
For training on ImageNet, we adopt the same data augmentation procedure as in the original Residual Networks (ResNets) paper~\citep{resnet}.

When training the layers from scratch, 
we found it useful to add a batch normalization layer~\citep{ioffe2015batch} before and after the TCL/TRL 
to avoid vanishing or exploding gradients,
and to make the layers more robust to changes 
in the initialization of the factors. 
In addition, we constrain the weights 
of the tensor regression 
by applying \(\ell_2\) normalization~\citep{salimans2016weight} 
to the factors of the Tucker decomposition. 

When experimenting with the tensor regression layer, 
instead of retraining the whole network each time 
it is possible to start from a pre-trained ResNet. 
We experimented with two settings: 
(i) We replaced the last average pooling, 
flattening and fully connected layers 
by either a TRL or a combination of TCL + TRL 
and trained these from scratch 
while keeping the rest of the network fixed;
and (ii) We investigate replacing the pooling and fully connected layers 
with a TRL that jointly learns the spatial pooling
as part of the tensor regression. 
In that setting, we also explore initializing the TRL 
by performing a Tucker decomposition 
on the weights of the fully connected layer.

\subsection{Large scale image classification}
First, we report results in the typical setting 
for large scale image classification on the the widely-used ImageNet-1K dataset, 
by learning the spatial pooling as part of the tensor regression. 

The ILSVRC dataset~\citep{imagenet} (ImageNet) 
is composed of \(1.2\) million images for training 
and \(50,000\) for validation, 
all labeled for 1,000 classes. 
Following~\citep{densenet,resnet,huang2016stochastic,he2016identity}, 
we report results on the validation set 
in terms of Top-1 accuracy and Top-5 accuracy across all \(1,000\) classes.
Specifically, we evaluate the classification error 
on single \(224 \times 224\) single center crop from the raw input images.

In this setting, we remove the average pooling layer 
and feed the tensor of size 
(batch size, number of channels, height, width) to the TRL, 
while imposing a rank of 1 on the spatial dimensions 
of the core tensor of the regression.
Effectively, this setting simultaneously learns weights 
for the multilinear spatial pooling as well as the regression.

\begin{table}
	\caption{Results obtained with a ResNet-101 architecture on ImageNet, learning spatial pooling as part of the TRL. We report the Top-1 and Top-5 accuracies, as well as the space savings obtained by replacing the fully connected layer with a tensor regression layer.
	Our approach enables large space savings with minimal impact on accuracy.
	In particular, for smaller space savings (about 25\%), our approach enables marginal improvements in performance, space savings of up to about 65\% do not impact performance; larger space savings translate into small decrease in performance.}\label{tab:imagenet}
      \centering
      \begin{tabular*}{0.9\textwidth}{@{\extracolsep{\fill}} llll}
        \toprule
        & \multicolumn{3}{c}{\textbf{Performance (\%)}} \\
        \cmidrule{2-4}
        \textbf{TRL rank}  &  \textbf{Top-1}  & \textbf{Top-5} & \textbf{Space savings} \\
        \midrule
        \emph{Baseline}  & \bf{77.1}  & \bf{93.4}  &   0   \\
        (300, 1, 1, 700) & \bf{77.2}  & \bf{93.5}  & 25.6  \\
        (200, 1, 1, 200) & \bf{77.1}  & \bf{93.2}  & 68.2  \\
        (120, 1, 1, 300) & \bf{76.7} & \bf{93.1}     & 71.2 \\
        (150, 1, 1, 150) & 76    & 92.9  & 76.6  \\
        (100, 1, 1, 100) & 74.6  & 91.7  & 84.6  \\
        (50 , 1, 1, 50)  & 73.6  & 91    & \bf{92.4}  \\
        \bottomrule
      \end{tabular*}
      \label{fig:table}
\end{table}

Our experimental results show that our method enables an
effective trade-off between performance and space savings
(Table~\ref{tab:imagenet}).
In particular, small space savings (e.g. about \(25\%\)) 
translate in  marginal increases in performance.
It is possible to obtain more than \(65\%\) space savings without impacting accuracy, 
and to reach larger performance space savings 
with litte impact on performance (e.g. almost \(80\%\) space savings 
with less than \(1\%\) decrease in Top-1 and Top-5 accuracy).
We can express the space savings of a model \(M\) 
with \(n_M\) total parameters in its fully connected layers 
with respect to a reference model \(R\) 
with \(n_R\) total parameters 
in its fully connected layers 
as \(1 - \frac{n_{M}}{n_{\text{R}}}\) (bias excluded).

To study the TRL in isolation, we consider the weights of a pre-trained model
and replace the flattening and fully connected layer with a TRL 
while keeping the rest of the network fixed.
In practice, to initialize the weights of the TRL in this setting, 
it is possible to consider the weights of the fully connected layer 
as a tensor of size 
(batch size, number of channels, 1, 1, number of classes) 
and apply a partial Tucker decomposition to it 
by keeping the first dimension (batch-size) untouched. 
The core and factors of the decomposition 
then give us the initialization of the TRL. 
The projection vectors over the spatial dimension 
are then initialized to 
\(\frac{1}{\text{height}}\) and\(\frac{1}{\text{width}}\), respectively. 
The Tucker decomposition was performed using TensorLy~\citep{tensorly}. 
In this setting, we show that we can drastically decrease 
the number of parameters with little impact on performance.

\subsection{Large scale phenotypical trait prediction from MRI data}

A major challenge in neuroscience and healthcare is analysing structure-rich data, 
such as 3D brain scans~\citep{mills2014methods,johnson2012brain}.
The human brain is a highly complex and structured organ.
It is composed of interconnected heterogeneous components, 
each of which is structurally distinct~\citep{sporns2005human}. 
The organizational properties are critical to the brain’s,
and the individual’s, overall health.
Studies of brain topology have revealed that structural changes 
are associated with cognitive function~\citep{andreasen1993intelligence} 
and diseases such as diabetes~\citep{raji2010brain}.
Retaining this topological information is thus critical 
in order to maximize modelling accuracy. 
However, flattening layers followed by fully connected layers 
discards that information and are therefore sub-optimal for the task. 
By contrast, our proposed tensor regression layers naturally leverages 
and preserves the topological information, allowing for better performance.
We empirically demonstrate this on the UK biobank MRI dataset, 
the largest imaging study of its kind~\citep{sudlow2015uk}, the results can be seen in Table~\ref{biobank}.

\begin{table}[ht]
  \caption{\textbf{Performance of for UK Biobank MRI} using a regular 3D-ResNet with a fully connected layer for prediction (\emph{baseline FC}) and with a 3D-ResNet where the fully connected layer was replaced with our proposed TRL. The ResNet models with TRL significantly outperforms the baseline version with a fully connected (FC) layer, as it preserves and leverages topological structure.}
  \label{biobank}
  \centering
  \resizebox{1\textwidth}{!}{
  \begin{tabular*}{1\textwidth}{@{\extracolsep{\fill}} llll}
    \toprule
    \multirow{2}{*}{\bf Method} & \multicolumn{1}{c}{\bf Age} 
                               & \multicolumn{1}{c}{\bf Gender} 
                               & \multicolumn{1}{c}{\bf BMI} 
    \\
    \cmidrule{2-4}
                            & MAE (years) & Classif.Err. (\%) & MAE (\(\mathrm{kg/m}^2\)) 
                            \\
    \midrule
    \textbf{baseline FC} & 2.96 & 0.79 & 2.37 
    \\
    (256, 3, 4, 3, 1)--TRL 
        & \bf{2.70} & \bf{0.53} & \bf{2.26} 
        \\ 
    (128, 3, 4, 3, 1)--TRL  & 2.72 & 0.60 & 2.33 
    \\
    (64, 2, 4, 2, 1)--TRL  & \bf{2.69} & 0.69 & 2.35 
    \\
    (32, 2, 4, 2, 1)--TRL & 2.78 & 0.73 & 2.38 
    \\
    \bottomrule
  \end{tabular*}
  }
\end{table}

We split the data into a training set containing \(11,500\) scans, 
a validation set of \(3,800\) scans and \(3,800\) scans for a held-out test set.
Each scan is a T1-weighted three-dimensional structural MRI with dimensions \(182\times218\times182\). 
We select three tasks: classifying gender, 
predicting body mass index (BMI) and predicting age from raw brain MRI scans. 
Predicting age is an important task as the difference 
between true and predicted age is a biomarker 
that has a number of clinical applications~\citep{cole2017predicting,kolenic2018obesity,franke2018premature}. Similarly, learning to predict BMI from MRI is a valuable objective, 
as influence of obesity on brain structure has already been established \cite{alosco2014body},
but its mechanism is not fully understood. 
For the third task of gender classification, 
differences between genders (more precisely biological sex) in brain functional mappings 
have already been described \citep{ingalhalikar2014sex}. 
The reasons and implications (if any) of these differences are not known,
but understanding the biological processes underpinning them 
could provide insights into how the brain works.
For gender prediction the classes are fairly balanced,
the male:female ratio in the UK Biobank is \(0.46:0.54\). 

We compare a standard ResNet with 3D convolutions (3D-ResNet) 
to a 3D-ResNet where the final fully connected layer was replaced with a TRL.
The baseline ResNet contains a global-average pooling layer 
that compresses all three spatial dimensions, 
giving an output size of \(512 \times n_{outputs} \). 
The models model were implemented using PyTorch~\citep{paszke2017automatic} 
and TensorLy~\citep{tensorly} and trained was done on a Tesla P100 GPU.
Both networks were trained from random initialization. 
The same number of iterations was used for all comparisons. 
After validating the number of iterations, 
we found the baseline and the TRL model required same number of iterations 
in order to reach convergence and the same number of iterations 
was subsequently used for all models and all experiments.

The results (table.~\ref{biobank}) demonstrate that our tensor-based architecture 
is able to learn a mapping between structural MRI and phenotypes. 
It performs significantly better on all tasks compared to a 3D-ResNet
with traditional average pooling operation and a fully connected layer.
In particular, the 3D-ResNet achieved a mean absolute error (MAE) of 2.96 years 
on the age-regression task compared with 2.70 years for our TRL.
Similarly, the TRL out-performed the baseline on the gender classifying task, 
achieving an error of only \(0.53 \%\) (accuracy of \(99.47 \%\)) 
compared to an error of \(0.79 \%\) for the baseline. 
For BMI regression, the TRL also performed with a notable improvement over the baseline. 
A secondary baseline, 3D-ResNet with no average pooling, 
failed to train at all due to unstable gradients.

These three important tasks demonstrate that the TRL is leveraging 
additional information in the structural MRI data beyond that of the baseline model. 
Our method improves on previously reported brain age accuracy
in UK Biobank~\citep{ning2018association}, and published studies~\citep{gaser2013brainage,kolenic2018obesity,cole2018brain} on other brain MRI datasets. 
The method also improves on BMI prediction beyond those 
previously published for UK Biobank data \citep{vakli2020predicting}.
Improved performance gives further support for use of machine learning 
as a support tool in neuroradiological research and clinical decision making.

One question is whether this improvements are due to the ability of the TRL 
to preserve and leverage topological structure or whether 
it is simply a result of the regularizing effect of the TRL. 
To answer this question, we experimented with a TRL,
where the entire \(6 \times 7 \times 6\) full-rank activation tensor 
was used for the age prediction task. 
Using the full-rank tensor the network achieved a MAE of 2.71 years,
which is similar to the results obtained with lower-rank set ups, 
and significantly better than the baseline. 
This empirically confirms the importance of leveraging topological structure.

\subsection{Ablation studies}

\paragraph{Synthetic setting}

To illustrate the effectiveness of the low-rank tensor regression, 
in terms of learning and data efficiency,
we first investigate it in isolation. 

We apply it to synthetic data 
\(y = \text{vec}(\mytensor{X}) \times \mymatrix{W} \) 
where each sample \(\mytensor{X} \in \myR^{(64 \time 64)}\) 
follows a Gaussian distribution \(\mathcal{N}(0, 3)\). 
\(\mymatrix{W}\) is a fixed matrix
and the labels are generated as 
\(y = \text{vec}(\mytensor{X}) \times \mymatrix{W}\). 
We then train the data on \(\mytensor{X} + \mytensor{E}\),
where \(\mytensor{E}\) is added Gaussian noise 
sampled from \(\mathcal{N}(0, 3)\). 
We compare i) a TRL with squared loss and 
ii) a fully connected layer with a squared loss. 

\begin{figure*}[h]
 	\centering
      \includegraphics[width=0.9\textwidth]{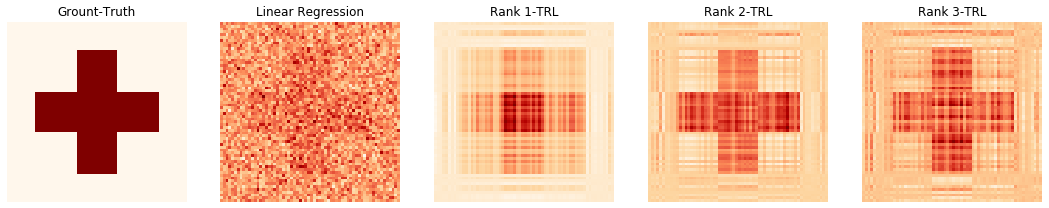}\\
      \includegraphics[width=0.9\textwidth]{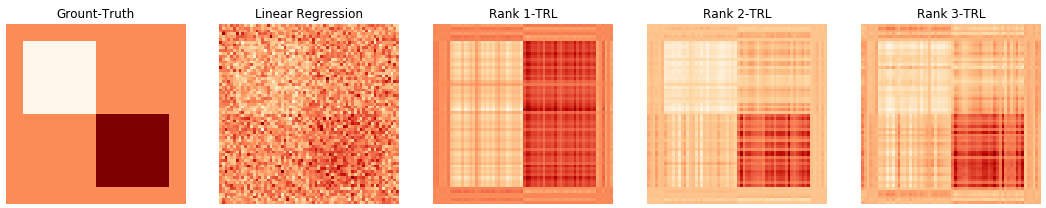}
    	\caption{Empirical comparison (\ref{fig:compare}) of the TRL against linear regression with a fully connected layer. 
 We plot the weight matrix of a TRL and a fully connected layer.
Due to its low-rank weights, 
the TRL better captures the structure in the weights 
and is more robust to noise.
}
    \label{fig:compare}
\end{figure*}
    
In Figure~\ref{fig:compare}, 
we show the trained weight of both a linear regression based on a fully connected layer 
and a TRL with various ranks, both obtained in the same setting. 
The TRL is able to better recover the ground-truth weight
due to the low-rank structure imposed on these, 
which allows to leverage the multi-linear structure
and acts as an implicit regularizer. 
This is in line with findings from existing works 
focusing on analytical solution to tensor regression problems~\cite{rabusseau2016low}. 
Additional regularizers such as Lasso (l1)~\cite{hoefling2010path,tibshirani2011solution}
can provide further advantages, especially in the face of noisy inputs. 
Going forward, we plan to investigate such additional regularization terms.

\begin{figure}[h]
    \centering
  	\includegraphics[width=0.55\linewidth]{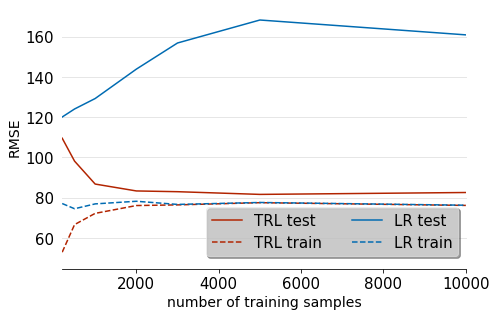}
  	\caption{Evolution of the RMSE as a function of the training set size for both the TRL and fully connected regression. Due to the low-rank structure of its weights tensor, the TRL requires less training data and is less prone to overfitting than fully connected layers.}
	\label{fig:synthetic-rmse}
\end{figure}

We also compare the data efficiency of a tensor regression layer 
compared to a linear regression, in the same setting,
by varying the number of training samples.
As can be observed in Figure \ref{fig:synthetic-rmse}, 
the TRL is easier to train on small datasets and less prone to overfitting, 
due to the low rank structure of its regression weights, as opposed to typical 
(fully connected) linear regression.

\paragraph{Impact of the tensor contraction layer}
We first investigate the effectiveness of the TCL 
using a VGG-19 network architecture~\citep{vgg}. 
This network is especially well-suited for our methods
because of its \(138,357,544\) parameters, 
\(119,545,856\) of which (more than 80\% of the total number of parameters) 
are contained in the fully-connected layers. 
By adding a TCL to contract the activation tensor 
prior to the fully connected layers, 
we can achieve large space saving. 

\begin{table}[h]
 \caption{Results obtained on ImageNet 
 by adding a TCL to a VGG-19 architecture. 
We reduce the number of hidden units proportionally
to the reduction in size of the activation tensor 
following the tensor contraction. 
Doing so allows more than 65\% space savings 
over all three fully connected layers 
(i.e. 99.8\% space saving over the fully connected layer replaced by the TCL) 
with no corresponding decrease in performance 
(comparing to the standard VGG network as a baseline).}
  \label{vgg}
  \centering
  \begin{tabular*}{0.9\textwidth}{@{\extracolsep{\fill}} lllll}
    \toprule
    \multicolumn{2}{c}{\bf Method}               &  \multicolumn{2}{c}{\textbf{Accuracy}}      & \multicolumn{1}{c}{\textbf{Space Savings}}   \\
    \cmidrule{1-2} \cmidrule{3-4}
    \textbf{TCL--size}         &  \textbf{Hidden Units}         &  \textbf{Top-1 (\%)}  & \textbf{Top-5 (\%)}         &   \multicolumn{1}{c}{\textbf{(\%)}}   \\
    \midrule
    \emph{Baseline}           &  4096                 &  68.7        &  88               &   0      \\
    (512, 7, 7)        &  4096                 &  \bf{69.4}  &  \bf{88.3}            &   -0.21  \\
    (384, 5, 5)        &  3072                 &  68.3        &  87.8             &   \textbf{65.87}  \\
    \bottomrule
  \end{tabular*}
\end{table}

Table~\ref{vgg} presents the accuracy obtained 
by the different combinations of TCL 
in terms of top-1 and top-5 accuracy as well as space saving. 
By adding a TCL that preserves the size of its input 
we are able to obtain slightly higher performance 
with little impact on the space saving (0.21\% of space loss) 
while by decreasing the size of the TCL we got more than \(65\%\)
space saving with almost no performance deterioration.

\paragraph{Overcomplete TRL}

\begin{table}[!ht]
  \caption{Results obtained with ResNet-50 on ImageNet. 
  The first row corresponds to the standard ResNet. 
  Rows 2 and 3 present the results obtained 
  by replacing the last average pooling, 
  flattening and fully connected layers with a TRL. 
  In the last row, we have also added a TCL.}
  \label{resnet}
  \centering
   \begin{tabular*}{0.9\textwidth}{@{\extracolsep{\fill}} c c c l l }
    \toprule
    \multicolumn{3}{c}{\bf Method} & \multicolumn{2}{c}{\bf Accuracy} \\
    \cmidrule{1-3}
    \textbf{Architecture} & \textbf{TCL--size} & \textbf{TRL rank} & \textbf{Top-1 (\%)}  & \textbf{Top-5 (\%)} \\
    \midrule
    ResNet-50 &  \multicolumn{2}{c}{NA (baseline)} &  74.58       &  92.06   \\
         ResNet-50      &  no TCL                 & (1000, 2048, 7, 7) &  73.6       &  91.3    \\
      ResNet-50         &  no TCL                 & (500, 1024, 3, 3) &  72.16        &  90.44    \\
    ResNet-50           &  (1024, 3, 3)      & (1000, 1024, 3, 3) &  73.43        &    91.3  \\
    ResNet-101 &  \multicolumn{2}{c}{NA (baseline)}   &  77.1       &  93.4   \\
  ResNet-101         & no TCL                 & (1000, 2048, 7, 7) &  76.45       &  92.9    \\
ResNet-101      &  no TCL                & (500, 1024, 3, 3) &  76.7        &  92.9    \\
ResNet-101       &  (1024, 3, 3)      & (1000, 1024, 3, 3) &  76.56        &    93  \\
    \bottomrule
  \end{tabular*}
\end{table}

We first tested the TRL with a ResNet-50 and a ResNet-101 architectures on ImageNet,
removing the average pooling layer 
to preserve the spatial information in the tensor.
The full activation tensor is directly passed on 
to a TRL which produces the outputs 
on which we apply softmax to get the final predictions. 
This results in more parameters as the spatial dimensions are preserved.
To reduce the computational burden 
but preserve the multi-dimensional information, 
we alternatively insert a TCL before the TRL. 

In Table \ref{resnet}, we present results obtained in this setting 
on ImageNet for various configurations of the network architecture. 
In each case, we report the size of the TCL 
(i.e. the dimension of the contracted tensor) 
and the rank of the TRL (i.e. the dimension of the core of the regression weights).

\paragraph{Choice of the rank of the TRL}
While the rank of the TRL 
is an additional parameter to validate, 
it turns out to be easy to tune in practice.

\begin{figure*}[!h]
    \centering
    \begin{subfigure}[b]{0.45\textwidth}
        \includegraphics[width=1\linewidth]{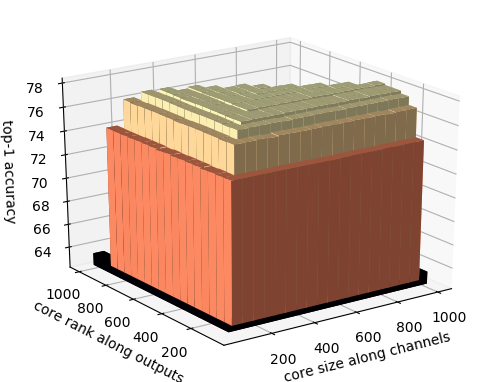}
        \caption{Accuracy as a function of the core size}
        \label{fig:barplot}
	\end{subfigure}
    \qquad
	\begin{subfigure}[b]{0.45\textwidth}
     	\includegraphics[width=1\linewidth]{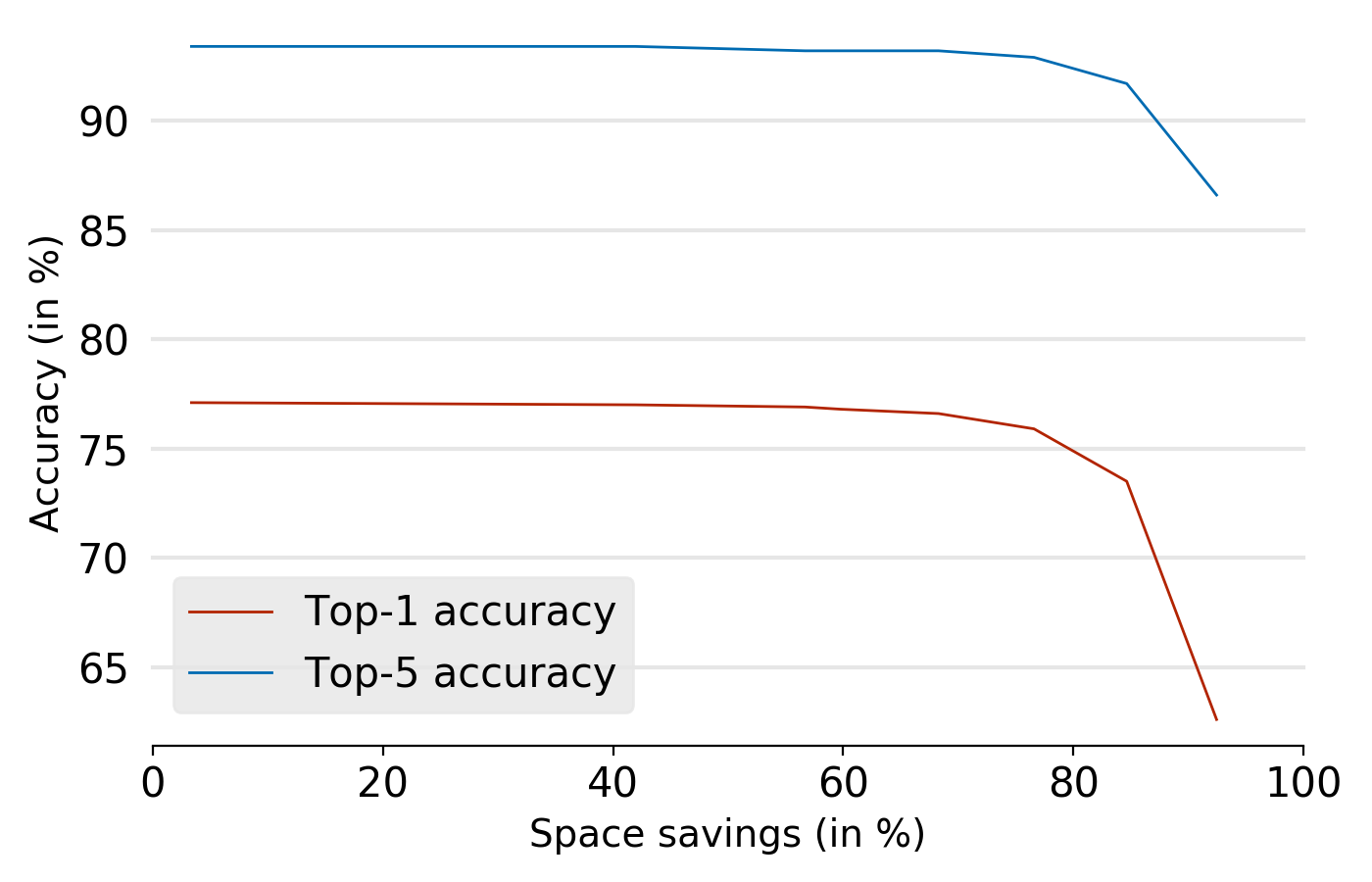}
    	\caption{Accuracy as a function of space savings}
    	\label{fig:curves}
	\end{subfigure}
     \caption{Study of the impact of the rank of the tensor regression layer on performance. On the left side, \ref{fig:barplot} shows the Top-1 accuracy (in \%)
as we vary the size of the core 
along the number of outputs 
and number of channels 
(the TRL does spatial pooling 
along the spatial dimensions, 
i.e., 
the core has rank 1 along these dimensions).
On the right side, \ref{fig:curves} shows the evolution of the Top-1 and Top-5 accuracy (in \%) 
as a function of the space savings by reducing the rank of the TRL (also in \%).
As can be observed, there is a large region for which
the reduction of the rank of the tensor regression layer
does not negatively impact the performance while enabling large space savings.
}
	\label{fig:resnetrank}
\end{figure*}

In Figure \ref{fig:resnetrank}, 
we show the effect on Top-1 and Top-5 accuracy 
of decreasing the size of the core tensor 
of the TRL. 
We also show the corresponding space savings.
The results suggest that choosing the rank is easy
because there is a large range 
of values of the rank 
for which the performance does not decrease. 
In particular, we can obtain up to \(80\)\% 
space savings with negligible impact on performance. 
 
\section{Conclusions}
Deep neural networks already operate on multilinear activation tensors, 
the structure of which is typically discarded 
by flattening operations and fully-connected layers. 
This paper proposed preserving and leveraging 
the tensor structure of the activations 
by introducing two new, end-to-end trainable, layers 
that enable substantial space savings 
while preserving and leveraging the multi-dimensional topological structure. 
The TCL that we propose reduces the dimension 
of the input without discarding its multi-linear structure, 
while TRLs directly map their input tensors
to the output with low-rank regression weights. 
These techniques are easy to plug in to existing architectures
and are trainable end-to-end.

Our experiments demonstrate 
that by imposing a low-rank constraint 
on the weights of the regression, 
we can learn a low-rank manifold 
on which both the data and the labels lie. 
Furthermore these new layers act as an additional type of regularization 
on the activations (TCL) and the regression weight tensors (TRL).
The result is a compact network 
that achieves similar accuracies with far fewer parameters.
The structure in the regression weight tensor 
allows for more interpretable models while requiring less data to train.
Going forward, we plan to apply the TCL and TRL to more network architectures 
and leverage recent work to avoid computational overhead 
from transpositions when computing tensor contractions.
 
\section*{Acknowledgements}
This research has been conducted using the UK Biobank Resource under Application Number 18545.
The authors would like to thank the editor and anonymous reviewers for the constructive feedback which helped improve this manuscript.

\end{document}